%% file: main.tex
\documentclass{article}
\usepackage{spconf,amsmath,amssymb,graphicx,cite}
\usepackage{algorithmic}
\usepackage{algorithm}
\usepackage{booktabs}

\usepackage{color}

\newcommand{\Hquad}{\hspace{0.0em}} 
\newcommand\mypar[1]{\noindent\textbf{#1}\Hquad}
\newcommand{\Sref}[1]{\S\ref{#1}}
\newcommand{\Fref}[1]{Figure~\ref{#1}}
\newcommand{\Tref}[1]{Table~\ref{#1}}
\newcommand{\argmax}{\mathop{\mathrm{argmax}}\limits}
\makeatletter
\def\blfootnote{\xdef\@thefnmark{}\@footnotetext}
\makeatother

\title{Align, Write, Re-order: Explainable End-to-End Speech Translation \\ via Operation Sequence Generation}

\name{Motoi Omachi$^{1*}$, Brian Yan$^{2*}$, Siddharth Dalmia$^2$, Yuya Fujita$^1$, Shinji Watanabe$^2$\thanks{$^{*}$Equal contribution}}
\address{
  $^1$Yahoo Japan Corporation, Tokyo, JAPAN; $^2$Carnegie Mellon University, PA, USA  }
\begin{document}
\ninept
\maketitle
\begin{abstract}

The black-box nature of end-to-end speech translation (E2E ST) systems makes it difficult to understand \textit{how} source language inputs are being mapped to the target language.
To solve this problem, we would like to simultaneously generate automatic speech recognition (ASR) and ST predictions such that each source language word is explicitly mapped to a target language word.
A major challenge arises from the fact that translation is a \textit{non-monotonic} sequence transduction task due to word ordering differences between languages -- this clashes with the \textit{monotonic} nature of ASR.
Therefore, we propose to generate ST tokens out-of-order while remembering how to re-order them later.
We achieve this by predicting a sequence of tuples consisting of a source word, the corresponding target words, and post-editing operations dictating the correct insertion points for the target word.
We examine two variants of such \textit{operation sequences} which enable generation of monotonic transcriptions and non-monotonic translations from the same speech input simultaneously.
We apply our approach to offline and real-time streaming models, demonstrating that we can provide \textit{explainable} translations without sacrificing quality or latency. 
In fact, the delayed re-ordering ability of our approach improves performance during streaming.
As an added benefit, our method performs ASR and ST simultaneously, making it faster than using two separate systems to perform these tasks. 

%We propose a Transformer-based sequence-to-sequence model predicting monotonic word alignment of transcriptions and translation to achieve automatic speech recognition (ASR) and speech translation (ST) simultaneously. Traditional pipeline systems, i.e., ASR followed by machine translation, cannot avoid the delay in obtaining translation because we must wait until ASR finishes. Recently proposed end-to-end (E2E) ST models can avoid such a delay, but it is hard to interpret translation errors because these models don't yield transcription. The proposed model runs faster than the pipeline systems by successively predicting the source, target word, and its position in the translation. Additionally, it is easy to interpret translation errors since our model yields transcriptions, which E2E ST models don't predict. Experimental results on English-to-German show that our model predicts high-quality transcriptions and translations.
\end{abstract}
\vspace{-2mm}
\section{Introduction}
\vspace{-2mm}

% introduction v2

Speech-to-text translation (ST) is an inherently compositional task consisting of first recognizing what was said and then translating the meaning.
Cascaded approaches to ST follow this two-staged processing by first performing automatic speech recognition (ASR) and then passing those predictions to a machine translation (MT) model \cite{waibel1996interactive, inaguma2021espnet, lam2021cascaded, zhang2022ustc}.
On the other hand, end-to-end (E2E) approaches obviate the need for the intermediate ASR task \cite{berard2016listen, bahar2019comparative, inaguma2021orthros, zhang2022revisiting} but recent works have demonstrated improved performance using E2E multi-staged architectures which reflect the compositional nature of ST \cite{sperber2019attention, bahar2021tight, dalmia2021searchable, yan-etal-2022-cmus}.
However, E2E systems still offer reduced explainability compared to cascaded systems which can use neural MT approaches that explicitly emit word alignment information \cite{stahlberg:2018}.

\begin{figure}
    \centering
    \begin{tabular}{c}
      \begin{minipage}{\linewidth}
      \begin{center}
      \includegraphics[width=.5\linewidth]{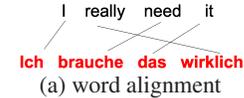} \\
      \vspace{-4mm}
      {(a) word alignment} \\
      \end{center}
      \end{minipage} \\
      \vspace{-2mm}
      \\
      \vspace{-2mm}
      \begin{minipage}{\linewidth}
      \begin{center}
      \includegraphics[width=.72\linewidth]{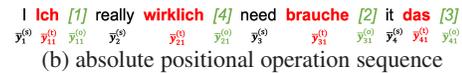} \\
      \vspace{-1mm}
      {(b) absolute positional operation sequence} \\
      \end{center}
      \end{minipage} \\
      \\
      \begin{minipage}{\linewidth}
      \begin{center}
      \includegraphics[width=.72\linewidth]{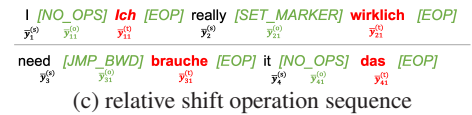} \\
      \vspace{-1mm}
      {(c) relative shift operation sequence} \\
      \end{center}
      \end{minipage} \\
    \end{tabular}
    \vspace{-3mm}
    \caption{Target sequences for our proposed E2E model ((b) and (c)) generated from source-\textcolor{red}{target} alignments (a). Symbols in brackets are post-editing operations dictating target word insertion order.}
    \vspace{-5mm}
    \label{fig: target sequence}
\end{figure}

One direction towards E2E ST explainability is to simultaneously transcribe and translate using the same model.
Prior works achieve this by interleaving chunks of ASR and ST predictions within one decoder \cite{dong2021consecutive, weller:2021} or by allowing dedicated decoders to model latent alignments between the tasks \cite{liu2020synchronous, le2020dual, yan2022ctc}. 
However, it remains difficult to obtain word-level hard-alignments between ASR and ST sequences from such methods.
Ideally, an explainable system would generate a source language word and its aligned target language word in tandem, but this would mean that the resultant translation sequence would be incorrectly ordered.
Recent MT works offer a potential solution for this re-ordering problem via non-monotonic generation methods capable of insertion-based decoding \cite{gu-etal-2019-insertion, emelianenko2019sequence, gois2020learning, ran2021guiding, xu2021editor}.
However, these highly flexible approaches are not directly compatible with the monotonic design of speech processing systems, particularly for real-time streaming \cite{pham:2019,ma:2019,Arivazhagan:2020,sperber:2020,chen:2021, IWSLT:2021, dong-etal-2022-learning}.

In this work, we seek to build explainable E2E ST models which 1) simultaneously generate ASR and ST predictions along with 2) word-level hard-alignment predictions while 3) avoiding degradation to offline and streaming translation quality.
In particular, we are interested in modeling \textit{insertions} to allow models to predict ASR tokens monotonically along with their corresponding ST tokens while remembering how to re-order the translation sequence later.

We propose to generate a sequence of tuples consisting of a source language word (for ASR), a corresponding target language word (for ST), and post-editing operations dictating the correct insertion point for the target word -- we refer to these as \textit{operation sequences} \cite{stahlberg:2018}.
Our first operation sequence predicts the absolute positions of each target word within the eventual translation sequence; however the learned absolute positions may not generalize.
To model insertion positions more relatively, we define a second operation sequence that models a shifting write header.
Our relative shift operation sequence is inspired by \cite{stahlberg:2018} which was proposed for MT, and we extend this approach to ST by also performing simultaneous prediction of source language words.
We demonstrate that our relative shift operation sequence enables explainable and high-quality E2E ST compared to baselines for offline and streaming settings with experiments on MuST-C \cite{di-gangi-etal-2019-must}. 
Further, we find that the delayed re-ordering ability of our approach improves the partial hypothesis during streaming.
Finally, an added benefit of our method is that simultaneous ASR/ST is faster than using two separate systems.
\vspace{-2mm}

\section{Background and Motivation}
\vspace{-2mm}

In this section, we review the black-box nature of direct approaches to E2E ST to motivate our explainable E2E ST approach in \Sref{ssec: prop}.
\vspace{-1mm}

\subsection{Direct E2E ST with ASR Multi-Tasking}
\label{ssec: e2e_st_overview}
The objective of the E2E ST is to directly predict a token sequence of the target $(t)$ translation ${\bf y}^{(t)}=\{y^{(t)}_l \in \mathcal{Y}^{(t)} \}_{l=1}^{L^{(t)}}$
% , which corresponds to token sequence of the source $(s)$ transcription ${\bf y}^{(s)}=\{y^{(s)}_l \in \mathcal{Y}^{(s)}\}_{l=1}^{L^{(s)}}$, 
from an input audio feature sequence ${\bf X}=\{x_{i} \in \Re^{D^{(x)}} \}_{i=1}^{I}$. $I$, $L^{(s)}$, and $L^{(t)}$ denote the sequence length of the input audio feature, transcription, and translation, respectively; $D^{(x)}$ denotes the dimension of input audio feature; and $\mathcal{Y}^{(t)}$ denotes the token vocabulary of target language.
To predict translation, a neural network (NN) is trained to maximize the following objective:
\begin{align}
    \label{eq: objective_e2eST}
    {\mathcal L}^{\mathrm (ST)}&=\log{p({\bf y}^{(t)}|{\bf X})} \nonumber \\
    &=\sum _{m=1} ^M \log p(y_m^{(t)}|y_{1:m-1}^{(t)}, {\bf X}),
\end{align}
which is frequently modeled using the attention-based encoder-decoder (AED) architecture~\cite{weiss:17,Karita:2019}. 
AED models trained to maximize Eq.~\eqref{eq: objective_e2eST} try to \textit{directly} predict a translation without knowledge of the source language, meaning it has to perform the complex mapping of a continuous speech input to a non-monotonic target language sequence.
To alleviate this challenge, we can consider that each audio feature sequence ${\bf X}$ has a corresponding token sequence of the source $(s)$ transcription ${\bf y}^{(s)}=\{y^{(s)}_l \in \mathcal{Y}^{(s)}\}_{l=1}^{L^{(s)}}$.
%To alleviate this challenge, we can consider that for each audio feature sequence ${\bf X}$ there is also a corresponding token sequence of the source $(s)$ transcription ${\bf y}^{(s)}=\{y^{(s)}_l \in \mathcal{Y}^{(s)}\}_{l=1}^{L^{(s)}}$. 
$\mathcal{Y}^{(s)}$ denotes the token vocabulary of source language. 
Therefore we apply a multi-tasked ASR and ST loss ${\mathcal L}$ as follows:
% the following multitask loss ${\mathcal L}$ to make optimization easier by considering the knowledge of ASR: 
\begin{align}
     \label{eq: objective_e2eASR}
    {\mathcal L}^{\mathrm (ASR)}&=\log{p({\bf y}^{(s)}|{\bf X})} \nonumber\\     & = \sum _{n=1} ^N \log p(y_n^{(s)}|y_{1:n-1}^{(s)}, {\bf X}). \\
     \label{eq: objective_multitask}
    {\mathcal L} &= \alpha {\mathcal L}^{(ST)} + (1-\alpha) {\mathcal L}^{(ASR)},
\end{align}
where $\alpha$ denotes a controllable interpolation (we use 0.3 in this study).
Using the NN trained with Eq.~\eqref{eq:  objective_multitask}, we can obtain hypothesized translation token sequence $\hat{\bf y}^{(t)}$ using the following decision:
\begin{align}
\hat{\bf y}^{(t)}=\argmax_{{\bf y}^{(t)} \in \mathcal{Y}^{(t)\ast} }{\log p({\bf y}^{(t)}|{\bf X})},\label{eq:decision}
\end{align}
%\begin{align}
%    \label{eq: obtain_translation}
%   \hat{\bf y}^{(t)}=\arg\max_{\bf{y}^(t) \in {\bf \mathcal{Y}}^{*(t)}}{\log p({\bf y}^{(t)}|{\bf X})},
%\end{align}
where $\mathcal{Y}^{(t)\ast}$ denotes all possible translation sequences.
Typically beam search approximates Eq.~\eqref{eq:decision}, during which the AED generates $m$-step output token $\hat{y}^{(t)}_{m}$ in a left-to-right autoregressive manner:
\begin{align}
   \label{eq: encoder}
   {\bf h} &= {\rm Encoder}(\bf X), \\
   \label{eq: s2s}
   \hat{y}^{(t)}_{m}&={\rm Decoder}(y^{(t)}_{1:m-1}, {\bf h}).
\end{align}
Similarly, an ASR decoder generates $n$-step source token $\hat{y}^{(s)}_{n}$, and ASR targets to feed the multi-tasked loss computation (Eq.~\eqref{eq:  objective_multitask}):
% To compute multitask loss (Eq.~\eqref{eq:  objective_multitask}), the following ASR decoder is used to compute token sequence of transcription:
\begin{align}
    \label{eq: asr_decoder}
   \hat{y}^{(s)}_{n}={\rm ASRDecoder}(y^{(s)}_{1:n-1},{\bf h}),
\end{align}
where both ST and ASR decoders share the same encoder.
Note that ST performance has been shown to benefit from encoder initialization with pre-trained ASR parameters (Eq.~\eqref{eq: objective_e2eASR}) ~\cite{Bansal:2019}.
% It is reported that ST performance is improved by replacing the Encoder (Eq.\eqref{eq: encoder}) and ASRDecoder (Eq.~\eqref{eq: asr_decoder}) with those pre-trained with ASR loss (Eq.~\eqref{eq: objective_e2eASR}) ~\cite{Bansal:2019}.

\subsection{Shortcomings of the Direct Approach}
\label{ssec:issues}
%\vspace{-1mm}

First and foremost, we are interested in improving the explainability of E2E ST.
Considering that the human process for speech-to-text translation entails first recognizing and then translating, the lack of an intermediate ASR output in the direct approach is a major divergence from our natural thought process.
Direct E2E ST models with ASR multi-tasking do generate ASR targets during training, but the ASR decoder (Eq.~\eqref{eq: asr_decoder}) is parallel to the ST decoder (Eq.~\eqref{eq: s2s}) -- in other words, ST predictions are not conditioned on any ASR predictions. Therefore, our \textit{first desideratum} is to simultaneously generate ASR and ST predictions such that each ST word may be explained by a corresponding ASR word.
Further our objective is not to simply build explainable ST models at all costs, but rather to enhance high-performance models with explainability -- this means that we cannot sacrifice translation quality or latency.
This latter consideration is particularly important for streaming models.
Therefore our \textit{second desideratum} is that our explainable models are as good as direct E2E ST toplines in terms of both translation quality and latency.

\vspace{-1mm}
\section{Proposed Framework}
\label{ssec: prop}
\vspace{-1mm}

We first propose to allow E2E ST models to predict word-aligned ASR and ST sequences simultaneously by formulating a general framework for generating target words out-of-order while remembering how to re-order them via post-editing to obtain final translations.
We then present two variants of such operation sequences which model absolute positions or relative shifts for target tokens to enable the aforementioned delayed re-ordering ability.

% This \textit{operation sequence} generation follows the monotonic nature of ASR, meaning that the target tokens are predicted first in an out-of-order fashion.
% We then use the predicted post-editing commands, which dictate insertion positions for each target token, to re-order the translation sequence.

\vspace{-1mm}

\subsection{Word-Aligned Simultaneous E2E ASR and ST}
\label{ssec: prop_model}
Suppose that word-level alignments between two languages are known.\footnote{In this work, we use the statistical tool MGIZA++~\cite{gao:2008}, which performs data-driven alignment without needing additional linguistic resources.}
Let us first re-formulate the ASR and ST targets as word sequences $\bar{\bf y}^{(s)}=\{\bar{\bf y}^{(s)}_{i}\}_{i=1}^{\bar{L}^{(s)}}$ and $\bar{\bf y}^{(t)}=\{\bar{\bf y}^{(t)}_{j}\}_{j=1}^{\bar{L}^{(t)}}$, where $\bar{L}^{(s)}$ and $\bar{L}^{(t)}$ denote the number of source and target words.
Note that we represent single words, $\bar{\bf y}_{i}^{(s)}$ or $\bar{\bf y}_{j}^{(t)}$, as a sequence of sub-word tokens.
We can then define word-aligned ASR/ST target sequences consisting of tuples of $(\bar{\bf y}^{(s)}_{i}, \{\bar{\bf y}_{ik}^{(t)}\}_{k=1}^{\bar{L}^{(t)}_{i}})$, where each source word $\bar{\bf y}^{(s)}_{i}$ is aligned to $\bar{L}^{(t)}_{i}$ target words in a one-to-many mapping.
Note that some source or target tokens may actually be un-aligned, so we must augment the respective vocabularies with special tokens [NO\_SRC] or [NO\_TGT].
Next, consider that simply concatenating all $i$-steps of target word chunks produces an out-of-order translation if there is any re-ordering between source and target languages (e.g., in \Fref{fig: target sequence}.a), thereby obscuring the intended meaning.

To resolve this out-of-order problem in word-aligned ASR/ST sequences, we propose augmenting each tuple with $\bar{L}^{(t)}_{i}$ post-editing operations $\{\bar{\bf y}_{ik}^{(o)}\}_{k=1}^{\bar{L}^{(t)}_{i}}$, obtaining an operation tuple:
\begin{align}
    {\bf a}_{i} = (\bar{\bf y}^{(s)}_{i}, \{\bar{\bf y}_{ik}^{(t)}, \bar{\bf y}_{ik}^{(o)}\}_{k=1}^{\bar{L}^{(t)}_{i}}).
\end{align}
These post-editing operations perform rule-based re-ordering of each aligned target word $\bar{\bf y}^{(t)}_{ik}$ by inserting them into correct positions, recovering the original order of the target translation -- we discuss particular rule sets to realize this function in the next section.

We can now define an \textit{operation sequence} ${\bf A}=\{{\bf a}_{i}\}_{i=1}^{M}$ and train an AED to predict ${\bf A}$ by replacing Eq.~\eqref{eq: objective_e2eST} with:
\begin{align}
  \label{eq: objective_align1}
  \mathcal{L}^{(OPS)} &= \log p({\bf A}|{\bf X}) \nonumber \\
  % &= \sum _{m=1} ^M \log p({\bf a}_{m} |, {\bf a}_{1:m-1}, {\bf X}) \nonumber \\  
  &=\sum _{i=1} ^{M^{\star}} \log p(y_i^{\star}|y_{1:i-1}^{\star}, {\bf X}),
\end{align}
where $y_i^{\star}$ denotes elements of the collapsed representation of ${\bf a}_{i}$ (i.e., words are collapsed into their token sequences) and $M^{\star}$ denotes the length of our target sequence.
Note that operation sequences are modeled autoregressively by first predicting the source, then the target, and finally the operation (Eq.~\eqref{eq: objective_align1}).
% Eq.~\eqref{eq: objective_align1} indicates that our model consider relations between ASR and ST and simultaneously predicts these sequences. 
On the other hand, the direct E2E ST model trained to maximize Eq.~\eqref{eq: objective_multitask} does not model any explicit relations between its parallel ASR and ST predictions.
Further note that we only replace the function of the main decoder Eq.~\eqref{eq: s2s} in our operation sequence AED models, meaning we do not need to sacrifice benefits of ASR multi-tasking (\Sref{ssec: e2e_st_overview}).

\subsection{Defining Post-Editing Operations}

\label{ssec: post-editing}
%We propose two types of sequences of the tuple ${\bf a}_{m}$, including source and target words, and post-editing command for indicating the target word position: absolute positional sequence (Fig.~\ref{fig: target sequence}(b)) and shift operation sequence (Fig.~\ref{fig: target sequence}(c)).
We examine two ways to represent the insertion position information of post-editing operations: the first uses absolute positions (e.g, \Fref{fig: target sequence}.b) and the second uses relative shifts (e.g., \Fref{fig: target sequence}.c).

\vspace{-3mm}
\subsubsection{Absolute Positional Operation Sequence}
\label{ssec: absolute}
% Let $\bar{\bf Y}_{m}^{(to)}=\{(\bar{\bf y}_{ik}^{(t)}, \bar{\bf y}_{ik}^{(o)})\}_{k=1}^{\bar{L}^{(t)}_{i}}$. 
We define absolute positional sequences with Backus-Naur form as: 
 % (BNF) as follows: 
{
% \vspace{-1mm}
\setlength{\leftmargini}{2mm}
\begin{itemize}
  \setlength{\parskip}{0mm} 
  \setlength{\itemsep}{0mm}
  \setlength{\labelsep}{0mm}
  \setlength{\itemindent}{0mm}
  \item[] $\textless{\bf A}\textgreater$ ::= $\textless{\bf A}\textgreater$$\textless{{\bf a}_{i}}^{*}\textgreater$[EOS]~$|$~$\textless{{\bf a}_{1}}^{*}\textgreater$[EOS]
  \item[] $\textless{{\bf a}_{i}}^{*}\textgreater$ ::= $\textless\bar{\bf y}^{(s)}_{i}\textgreater$[BL]$\textless{\bf T}\textgreater$
  \item[] $\textless{\bf T}\textgreater$ ::= $\textless{\bf T}\textgreater$$\textless{\bar{\bf y}^{(t)}_{ik}}\textgreater$$\textless{\bar{\bf y}^{(o)}_{ik}}\textgreater$~$|$~$\textless{\bar{\bf y}^{(t)}_{i1}}\textgreater$$\textless{\bar{\bf y}^{(o)}_{i1}}\textgreater$
  \item[] $\textless\bar{\bf y}^{(s)}_{i}\textgreater$ ::= $\bar{\bf y}^{(s)}_{i}$~$|$~[NO\_SRC]
  \item[] $\textless\bar{\bf y}^{(t)}_{ik}\textgreater$ ::= $\bar{\bf y}^{(t)}_{ik}$~$|$~[NO\_TGT]
  \item[] $\textless\bar{\bf y}^{(o)}_{ik}\textgreater$ := [$n$]~$|$~[$-1$],
\end{itemize}
% \vspace{-1mm}
}
\noindent
where $::=$ and $|$ denote definition and choice, respectively; [BL] and [EOS] denote the blank symbol for separating source and target words and the end of the sequence, respectively; 
%$\bar{\bf Y}_{i}^{(to)}$ denotes the tuples $\{(\bar{\bf y}_{ik}^{(t)}, \bar{\bf y}_{ik}^{(o)})\}_{k=1}^{\bar{L}^{(t)}_{i}}$; 
$\bar{\bf y}^{(s)}_{i}$, $\bar{\bf y}^{(t)}_{ik}$ denotes $i$-th source word in the transcription and $k$-th target word which corresponds to $\bar{\bf y}^{(s)}_{i}$; 
%and $[n], which is the terminal symbol of $\textless\bar{\bf y}^{(o)}_{ik}\textgreater$, (where $n \in \mathbb{N} $)
and $[n]$ (where $n \in \mathbb{N} $) denotes the position of $\bar{\bf y}^{(t)}_{ik}$ in the translation. 
We set $n$ as $-1$ when $\bar{\bf y}^{(t)}_{ik}$ is [NO\_TGT]. %[BL] is used to separate source and target words.

To restore the transcription and translations, we prepare sufficient-length buffers for the source transcription $\mathcal{S}$ and the target translation $\mathcal{T}$. Note that all elements of the $\mathcal{S}$ and $\mathcal{T}$ are initialized with [NO\_SRC] or [NO\_TGT]. 
During run-time, our model outputs the absolute positional sequence tokens, and these tokens are inserted into FIFO queue $\mathcal{Q}$. When the model outputs the post-editing command token $\textless\bar{\bf y}^{(o)}_{ki}\textgreater$, we obtain sequences of $\textless\bar{\bf y}^{(s)}_{i}\textgreater$[BL]$\textless\bar{\bf y}^{(t)}_{i}\textgreater$$\textless\bar{\bf y}^{(o)}_{ki}\textgreater$ or $\textless\bar{\bf y}^{(t)}_{i}\textgreater$$\textless\bar{\bf y}^{(o)}_{ki}\textgreater$ by dequeuing $\mathcal{Q}$. 
Then the $i$-th element of $\mathcal{S}$ is rewritten with $\bar{\bf y}^{(s)}_{i}$ when the dequeued sequence includes source word.
And the $\textless\bar{\bf y}^{(o)}_{ki}\textgreater$-th element of $\mathcal{T}$ is rewritten with $\textless\bar{\bf y}^{(t)}_{ki}\textgreater$. If $\textless\bar{\bf y}^{(o)}_{ki}\textgreater$ indicates $-1$, we skip this step. This step is repeated until the end of the sequence, and we obtain the source transcription and the target translations from $\mathcal{S}$ and $\mathcal{T}$.

% The proposed absolute positional sequence helps us to interpret the generating process of transcription and translation. However, the learned absolute positions may not generalize.

\subsubsection{Relative Shift Operation Sequences}
\label{ssec: operation}
%%%%%%%%%%%%%%%
\begin{figure}
    \centering
    \includegraphics[width=.75\linewidth]{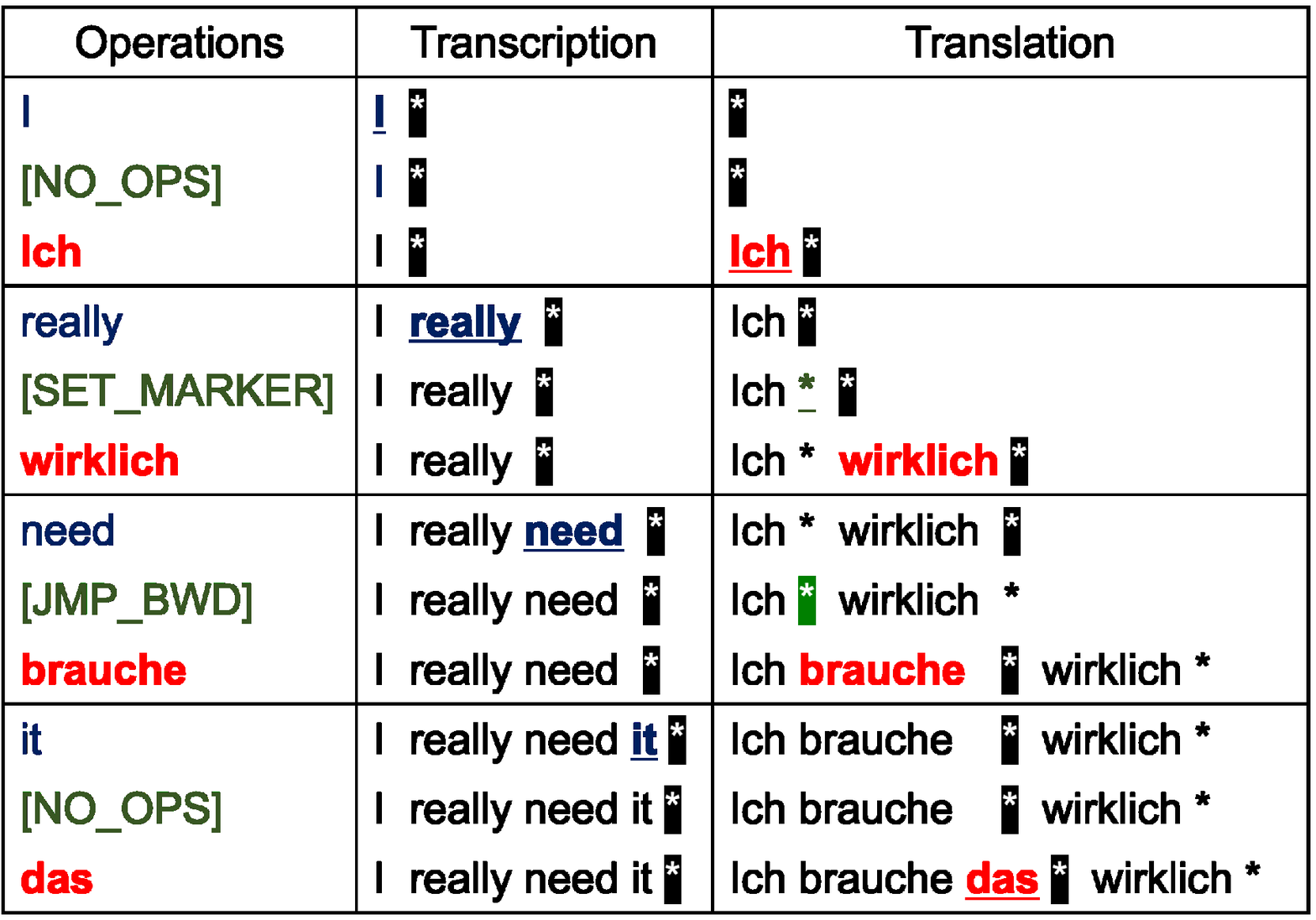}
    \vspace{-3mm}
    \caption{Restoring transcription and translation from the relative shift operation sequence.  [EOP] operations are omitted for space. * denotes a marker and positions of the write-heads are highlighted.}
    \label{fig: example_restoring}
    \vspace{-1mm}
\end{figure}
%%%%%%%%%%%%%%%
\cite{stahlberg:2018} originally proposed an explainable MT model which models insertions using a moving write header -- we adapt this approach for simultaneous ASR/ST.
% where source tokens are mapped to predicted target tokens -- we adapt this approach to ST, where both source are target tokens are predictions.
In \cite{stahlberg:2018}, the operation sequence is defined as the stack operations of the read-head ${p}_{\mathrm r}^{(s)}$ for the transcription $\mathcal{S}$ and the write-head ${p}_{\mathrm w}^{(t)}$ for translation $\mathcal{T}$ using following operations:
{
% \vspace{-1mm}
\setlength{\leftmargini}{3mm}
\begin{itemize}
  \setlength{\parskip}{0cm} 
  \setlength{\itemsep}{0cm}
  \item[]{\bf POP\_SRC} moves ${p}_{\mathrm r}^{(s)}$ right by one token
  \item[] {\bf SET\_MARKER (SM)} inserts a marker symbol into $\mathcal{T}$ at ${p}_{\mathrm w}^{(t)}$
    % \item[] {\bf SET\_MARKER (SM)} inserts a marker symbol into $\mathcal{T}$ at the position ${p}_{\mathrm w}^{(t)}$.

  \item[] {\bf JMP\_FWD (JF)} moves ${p}_{\mathrm w}^{(t)}$ to the closest left maker position
  \item[] {\bf JMP\_BWD (JB)} moves ${p}_{\mathrm w}^{(t)}$ to the closest right maker position
  \item[] {\bf INSERT($t$)} inserts a target token $t$ into  $\mathcal{T}$ at the position ${p}_{\mathrm w}^{(t)}$
\end{itemize}
% \vspace{-1mm}
}
\noindent
Since we do not know the transcription in advance, we remove ${p}_{\mathrm r}^{(s)}$ and {\bf POP\_SRC}. Instead, we define write-head $p_{\mathrm w}^{(s)}$ for transcription $\mathcal{S}$.
We also revise {\bf INSERT} into as follows:
{
% \vspace{-1mm}
\setlength{\leftmargini}{3mm}
\begin{itemize}
  \setlength{\parskip}{0cm} 
  \setlength{\itemsep}{0cm}
  \item[] {\bf INSERT-t($t$)} inserts a target token $t$ into $\mathcal{T}$ at ${p}_{\mathrm w}^{(t)}$
  \item[] {\bf INSERT-s($s$)} inserts a src tok $s$ into $\mathcal{S}$ at ${p}_{\mathrm w}^{(s)}$ and moves ${p}_{\mathrm w}^{(s)}$ right
  % \item[]\hphantom{000000000000}and moves ${p}_{\mathrm w}^{(s)}$ right by one token.
\end{itemize}
% \vspace{-1mm}
}

%Note that we don't need to set operations {\bf SET\_MARKER}, {\bf JMP\_FWD}, and {\bf JMP\_BWD} for ${p}_{\mathrm w}^{(s)}$ because our model output the source words monotonically. 
Our relative shift operation sequences are obtained in the same manner as \cite{stahlberg:2018} and are defined with Backus-Naur form as:
%\begin{align*}
%&<{\bf A}> ::= <{\bf A}><{{\bf a}_{i}}^{*}>\mathrm{[EOS]}~|~<{{\bf a}_{1}}^{*}>\mathrm{[EOS]} \\
%&<{{\bf a}_{i}}^{*}> ::= <\bar{\bf y}^{(s)}_{i}><\bar{\bf Y}_{i}^{(to)}> \mathrm{[EOP]} \\
%&<\bar{\bf Y}_{i}^{(to)}> ::= <\bar{\bf Y}_{i}^{(to)}><\bar{\bf y}^{(t)}_{i1}><{\bar{\bf y}^{(o)}_{i1}}>~|~<{\bar{\bf y}^{(t)}_{i1}}><{\bar{\bf y}^{(o)}_{i1}}> \\
%&<\bar{\bf y}^{(s)}_{i}> ::= \bar{\bf y}^{(s)}_{i}~|~\mathrm{[NO\_SRC]} \\
%&<\bar{\bf y}^{(t)}_{ik}> ::= \bar{\bf y}^{(t)}_{ik}~|~\mathrm{[NO\_TGT]} \\
%&<\bar{\bf y}^{(o)}_{ik}> ::= <\bar{\bf y}^{(o)}_{ik}><\bar{y}^{(o)}_{ikj}>~|~<\bar{y}^{(o)}_{ik1}> \\
%&<\bar{y}^{(o)}_{ikj}> :: = \mathrm{[JF]}~|~\mathrm{[JB]}~|~\mathrm{[SM]}~|~\mathrm{[NO\_OPS]}
%\end{align*}
{
% \vspace{-1mm}
\setlength{\leftmargini}{2mm}
\begin{itemize}
  \setlength{\parskip}{0cm} 
  \setlength{\itemsep}{0cm}
  \setlength{\labelsep}{0mm}
  \setlength{\itemindent}{0mm}
  \item[] $\textless{\bf A}\textgreater$ ::= $\textless{\bf A}\textgreater$$\textless{{\bf a}_{i}}^{*}\textgreater$[EOS]~$|$~$\textless{{\bf a}_{1}}^{*}\textgreater$[EOS]
  \item[] $\textless{{\bf a}_{i}}^{*}\textgreater$ ::= $\textless\bar{\bf y}^{(s)}_{i}\textgreater$$\textless{\bf T}\textgreater$[EOP]
  \item[] $\textless{\bf T}\textgreater$ ::= $\textless{\bf T}\textgreater$$\textless{\bar{\bf y}^{(o)}_{ik}}\textgreater$$\textless{\bar{\bf y}^{(t)}_{ik}}\textgreater$~$|$~$\textless{\bar{\bf y}^{(o)}_{11}}\textgreater$$\textless{\bar{\bf y}^{(t)}_{11}}\textgreater$
  \item[] $\textless\bar{\bf y}^{(s)}_{i}\textgreater$ ::= $\bar{\bf y}^{(s)}_{i}$~$|$~[NO\_SRC]
  \item[] $\textless\bar{\bf y}^{(t)}_{ik}\textgreater$ ::= $\bar{\bf y}^{(t)}_{ik}$~$|$~[NO\_TGT]
  \item[] $\textless\bar{\bf y}^{(o)}_{ik}\textgreater$ ::= $\textless\bar{\bf y}^{(o)}_{ik}\textgreater$$\textless\bar{y}^{(o)}_{ikj}\textgreater$~$|$~$\textless\bar{y}^{(o)}_{ikj}\textgreater$
  \item[] $\textless\bar{y}^{(o)}_{ikj}\textgreater$ ::= [JF]~$|$~[JB]~$|$~[SM]~$|$~[NO\_OPS],
\end{itemize}
% \vspace{-1mm}
}
\noindent
where [EOP] and [NO\_OPS] denote the end of operation sequence for $\textless\bar{\bf y}^{(s)}_{i}\textgreater$ and no operation is required for $\textless\bar{\bf y}^{(t)}_{ik}\textgreater$, respectively.

Similar to the absolute positional sequence, the relative shift operation sequence is inserted into the FIFO queue $\mathcal{Q}$ during run-time. When the model outputs [EOP], we obtain $\textless{\bf a}^{*}_{i}\textgreater$ by dequeuing $\mathcal{Q}$. Then, we can update $\mathcal{S}$ and $\mathcal{T}$ based on the operations. This operation is repeated until the end of the sequence, resulting in transcription and translations from $\mathcal{S}$ and $\mathcal{T}$ by removing the marker symbols.
Figure \ref{fig: example_restoring} depicts how the transcriptions and translations are restored from the proposed relative shift operation sequence.

% The shift operation sequence generates translations and transcriptions monotonically. This feature is suitable for the case that the source and target languages are syntactically different, i.e., the sentence pairs required to consider reordering.

    % Results comparing the ASR and ST performances, measured by $\%$ WER and BLEU, of our proposed simultaneous ASR/ST methods vs. baselines consisting of separate ASR and ST.
    % The 1st horizontal partition shows offline models while the 2nd shows streaming models -- latancy is measured via Average Lagging (AL) \cite{ma2020simulmt}.
    % Results are on tst-COMMON of MuST-C English$\rightarrow$German and English$\rightarrow$French \cite{di-gangi-etal-2019-must}.
\begin{table*}[t]
  \centering
    \caption{
    Comparison of simultaneous E2E ASR/ST models using \textbf{absolute position} vs. \textbf{relative shift operation} sequences across offline and streaming settings. 
    Better results between these two methods are \textbf{bolded} and any simultaneous results which reach/surpass the topline results of single-task E2E models are further \underline{\textbf{underlined}}.
    ASR and ST performances, as measured by $\%$ WER and BLEU, are shown on tst-COMMON / tst-HE  of MuST-C English$\rightarrow$German and English$\rightarrow$French \cite{di-gangi-etal-2019-must}.
    Average Lagging (AL) \cite{ma2020simulmt} is shown for streaming models.
    }
    \vspace{-2mm}
    \include{tables/main2}
    \label{tab:main}
%    \vspace{-5mm}
\end{table*}

\section{Results}
\subsection{Experimental setup}
\mypar{Data:} We conduct experiments using the English-German and English-French pairs of the MuST-C corpus \cite{di-gangi-etal-2019-must}. We apply the tokenizer of the Moses toolkit~\cite{koehn2007moses} for training sentences. Then we remove the samples whose length ratio between the source and target sequences is greater than five or the target sequence length is longer than 150 tokens. To compute the word alignment, we use MGIZA++~\cite{gao:2008}. For training MGIZA++ model, we set the number of iterations to 5 for HMM, Model 1, Model2, Model 3, and 10 for Model 4; and we use the deficient distortion model for the empty word to reduce the mapping of the target word into [NO\_SRC].

\mypar{Models:} Models are trained using ESPnet \cite{watanabe2018espnet, inaguma-etal-2020-espnet}.
We use $4000$-vocabulary and $16000$-vocabulary BPE \cite{sennrich2015neural} units for source transcription and target \text{operation sequence}.
We also use $16000$-vocabulary BPE units for direct E2E toplines for a fair comparison.
ASR and ST models both use conformer encoders \cite{gulati:2020, guo2021recent} with $12$ blocks, $4$ heads, $31$ kernel size, $2048$ feed-forward dim, and $256$ attention dim.
We initialize the ST encoder with pre-trained ASR parameters for faster convergence.
ASR and ST attentional decoders consist of $6$ blocks, $4$ heads, and $2048$ feed-forward dim.
Streaming models follow the blockwise method in \cite{tsunoo2021streaming} with $40$ block size, $16$ hop size, and $16$ look-ahead.
All models are trained for $40$ epochs and decoded using $10$ beam and $2048$ sim chunk length for streaming with repeat detection \cite{tsunoo2021streaming}.

\mypar{Evaluation:} We evaluate ASR via word error rate (WER$\downarrow$) and ST via de-tokenized case-insensitive BLEU($\uparrow$) \cite{post-2018-call}. We evaluate streaming latency via Average Lagging (AL$\downarrow$) \cite{ma2020simulmt}; for separate ASR/ST baselines, we consider the latency of both tasks together.

\subsection{Results and Discussion}
%------------------
\Tref{tab:main} presents our main results on two language pairs across offline and streaming settings.
We found that the relative shift operation sequence produces better translations than absolute position operation sequences for both offline (\texttt{B1} vs. \texttt{B2}) and streaming models (\texttt{B3} vs. \texttt{B4}).
The absolute positions appear to inhibit generalization, and we found that these models skewed towards overly short hypotheses even when brevity penalty was applied during beam search.
On the other hand, the relative position information from the relative shift operation sequence appears to avoid these pitfalls.
These explainable models achieve comparable performance compared to topline E2E ST models which solely optimize toward translation quality in the offline setting (\texttt{A2} vs. \texttt{B2}).
Impressively, in the streaming setting, our operation sequence models even surpass the performance of dedicated toplines without additional latency (\texttt{A4} vs. \texttt{B4}), suggesting that the ability to delay re-ordering is particularly useful for streaming ST.
This improvement is particularly noticeable in the English-German tst-COMMON set, where the relative shift operation sequence yields $0.9$ BLEU gain; we found that this particular set required the most re-ordering between source and target languages.

Since our ST models are also simultaneously predicting ASR transcriptions, we examine the ASR quality compared to topline models which solely focus on ASR.
In \Tref{tab:main}, it is clear that the burden of producing explainable ST predictions hinders our operation sequence models from matching the topline ASR models -- operation sequence models must not only transcribe, but also do so in a way that explains the translations.
We also noted that decoding longer translations caused the simultaneous transcriptions to degrade in WER; we do not seek to optimize the ASR side of this trade-off in this work.
Nonetheless, the simultaneous transcriptions are reasonable outputs which enable our desired word-level explainability.
Further, these transcriptions come with no additional latency, so our simultaneous ASR/ST models are faster than two separate dedicated models performing both tasks.

% Our simultaneous ASR/ST model with absolute position sequences (\Sref{ssec: absolute}) exhibits degraded performance compared to separate ASR/ST baselines (\texttt{A1} vs. \texttt{A2} and \texttt{B1} vs. \texttt{B2}). 
% Our operation sequence models outperformed ST baselines for En-De by 0.4 in offline (\texttt{A1} vs. \texttt{A3}) and 0.9 in streaming settings (\texttt{B1} vs. \texttt{B3}), suggesting that the re-ordering capability assisted the generation of German words which frequently map non-monotonically to English~\cite{navratil2012comparison}. 
% The operation sequence models also maintained within a $0.1$ range of the ST baselines for En-Fr where less input-to-output re-ordering is expected.
% Further, the AL of our simultaneous ASR/ST models exhibit a large streaming advantage as unlike separate ASR/ST setups which have to perform two decodings in tandem while our method simply requires one operations sequence decoding to obtain both ASR and ST predictions. 
% However, one drawback of our method is a distinct drop in ASR performance for both offline and streaming -- this is likely due to the increased complexity of the generation task, which must additionally predict translation and operation targets.

% \section{Discussion}

\begin{figure}
\centering
\includegraphics[width=0.85\linewidth]{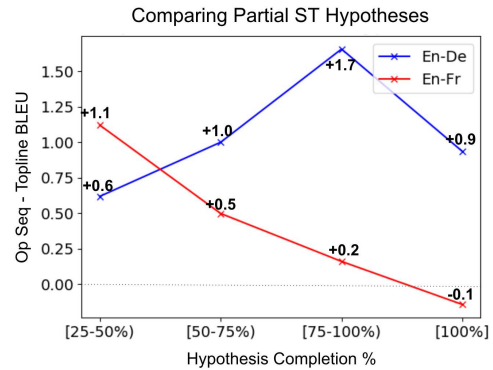}
\vspace{-3mm}
\caption{Comparison of partially generated ST hypotheses during streaming between relative shift operation sequence models vs. direct ST toplines, measured by absolute BLEU difference on dev sets.}
\vspace{-2mm}
\label{fig:partial}
\end{figure}

Finally, we are interested in the re-ordering capability of our operational sequence methods during streaming, where systems need to generate translations while still reading the speech signal.
We can apply operation sequences \textit{on-the-fly} instead of during post-editing.
This allows our systems to produce target words according to the order that they are spoken (monotonically) instead of waiting until all of the preceding target words have been produced (as in direct E2E systems).
% We therefore, examine the quality of partial hypotheses from our streaming operation sequence model compared to baselines for both language pairs.
As shown in \Fref{fig:partial}, relative shift operation sequence models consistently generate better partial hypotheses during streaming than topline direct models.
For instance in the En-Fr case, this advantage is most apparent for very early hypotheses and the gap between the models closes as hypotheses reach their full lengths.
% suggesting that on-the-fly operations improve the intelligibility of translations throughout streaming ST.
% In fact, the full hypotheses of the baseline En-Fr are slightly better by 0.1 BLEU, but the operation sequence model still produces better partial hypotheses, demonstrating the great impact that re-ordering has on translation quality.

\section{conclusion}
We propose to build E2E models which simultaneously perform ASR and ST by replacing targets with operation sequences which describe word-level alignments and post-editing commands for re-ordering target language words which were produced out-of-order.
Our models using relative shift operation sequences achieve explainability without sacrificing translation quality or achievement.
Further, these models actually outperformed direct E2E ST toplines in streaming settings due to an improved word re-ordering ability.\blfootnote{Authors acknowledge Graham Neubig's invaluable contributions.}

% We proposed a model producing source and target words and post-editing operations for reordering to achieve simultaneous E2E ASR and ST. 
% The proposed model outputs better translations during run-time while the transcription quality needs to be improved. Our model has a trade-off between the ASR and ST performance. In the future, we will tackle solving such trade-offs.

% \section{acknowledgements}

\clearpage
\footnotesize
\bibliographystyle{IEEEbib}
\bibliography{strings,refs}

\end{document}

%% file: tables/main2.tex
\resizebox {0.95\linewidth} {!} {
\begin{tabular}{cl|c|c|ccc|ccc}
\toprule
& & & & \multicolumn{3}{c|}{\underline{\textsc{En$\rightarrow$De}}} & \multicolumn{3}{c}{\underline{\textsc{En$\rightarrow$Fr}}}\\
\texttt{ID} & Model & Type &  \# Params & WER($\downarrow$) & BLEU($\uparrow$) & AL($\downarrow$) & WER($\downarrow$) & BLEU($\uparrow$) & AL($\downarrow$) \\
\midrule
\texttt{A1} & Direct E2E ASR \cite{watanabe2018espnet} (\textit{Topline}) & Offline & 45M & 7.7 / 6.7 & \textit{no ST} & \textit{n/a} & 9.9 / 8.1 & \textit{no ST} & \textit{na} \\ 
\texttt{A2} & Direct E2E ST \cite{inaguma-etal-2020-espnet} (\textit{Topline}) & Offline & 63M & \textit{no ASR} & 24.8 / 22.9 & \textit{n/a} & \textit{no ASR} & 35.9 / 32.9 & \textit{n/a} \\ 
\midrule
\texttt{A3} & Direct E2E ASR \cite{watanabe2018espnet} (\textit{Topline}) & Streaming & 45M & 12.0 / 9.9 & \textit{no ST} & 3365 & 17.9 / 14.2 & \textit{no ST} & 3279 \\
\texttt{A4} & Direct E2E ST \cite{inaguma-etal-2020-espnet} (\textit{Topline}) & Streaming & 63M & \textit{no ASR} & 22.2 / 20.1 & 5750 & \textit{no ASR} & 32.1 / 29.5 & 5648 \\
\midrule
\midrule
\texttt{B1} & Simul. E2E ASR/ST w/ Absolute Position & Offline & 63M & 12.1 / 12.0 & 20.4 / 17.9 & \textit{n/a} & 14.8 / 16.2 & 28.7 / 26.0 & \textit{n/a} \\
\texttt{B2} & Simul. E2E ASR/ST w/ Relative Shift & Offline & 63M & \textbf{10.1} / \textbf{9.9} & \underline{\textbf{25.2}} / \textbf{22.1} & \textit{n/a} & \textbf{12.5} / \textbf{11.3} & \textbf{35.8} / \underline{\textbf{32.9}} & \textit{n/a} \\
\midrule
\texttt{B3} & Simul. E2E ASR/ST w/ Absolute Position & Streaming & 63M & 18.7 / 17.2  & 18.4 / 16.5 & 5795 & 25.3 / 23.4 & 28.7 / 24.5 & 5684 \\
\texttt{B4} & Simul. E2E ASR/ST w/ Relative Shift & Streaming & 63M & \textbf{16.1} / \textbf{14.9} & \underline{\textbf{23.1}} / \underline{\textbf{20.1}} & 5786 & \textbf{19.3} / \textbf{16.9} & \underline{\textbf{32.1}} / \underline{\textbf{29.6}} & 5770 \\
\bottomrule
\end{tabular}
}